\documentclass[a4paper]{article}
\pdfoutput=1

\usepackage{scrextend}
\usepackage{CJKutf8}
\usepackage{graphicx}
\usepackage{float}
\usepackage{subfig}
\usepackage[table]{xcolor}
\usepackage{hyperref}
\usepackage{amsmath}
\usepackage{INTERSPEECH2021}

\title{g2pW: A Conditional Weighted Softmax BERT \\
for Polyphone Disambiguation in Mandarin}
\name{Yi-Chang Chen${^1}$\qquad
Yu-Chuan Chang${^1}$\qquad
Yen-Cheng Chang${^1}$\qquad
Yi-Ren Yeh${^2}$}
\address{
  $^1$E.SUN Financial Holding CO., LTD., Taiwan\\
 $^2$Department of Mathematics, National Kaohsiung Normal University, Taiwan
}
\email{
\{ycchen-20839, steven-20841, ycchang-21549\}@esunbank.com.tw, yryeh@nknu.edu.tw
}

\begin{document}

\maketitle
\begin{abstract}

Polyphone disambiguation is the most crucial task in Mandarin grapheme-to-phoneme (g2p) conversion. Previous studies have approached this problem using pre-trained language models, restricted output, and extra information from Part-Of-Speech (POS) tagging. Inspired by these strategies, we propose a novel approach, called g2pW, which adapts learnable softmax-weights to condition the outputs of BERT with the polyphonic character of interest and its POS tagging. Rather than using the hard mask as in previous works, our experiments show that learning a soft-weighting function for the candidate phonemes benefits performance. In addition, our proposed g2pW does not require extra pre-trained POS tagging models while using POS tags as auxiliary features since we train the POS tagging model simultaneously with the unified encoder. Experimental results show that our g2pW outperforms existing methods on the public CPP dataset. All codes, model weights, and a user-friendly package are publicly available.



\end{abstract}
\noindent\textbf{Index Terms}: polyphone disambiguation, grapheme to phoneme, weighted softmax, BERT

\section{Introduction}
\label{intro}


Mandarin grapheme-to-phoneme (G2P), which converts Chinese texts into pronunciation (Bopomofo or Pinyin), is a crucial component of Mandarin text-to-speech (TTS) systems. In Mandarin G2P conversion, the most important task is to distinguish the pronunciation of a polyphonic character, called polyphone disambiguation. Specifically, polyphone disambiguation aims to identify the correct pronunciation of the given polyphonic characters within a sentence. According to previous studies, polyphone disambiguation approaches typically can be divided into rule-based and learning-based approaches.


\begin{CJK*}{UTF8}{bsmi}
The rule-based approaches \cite{Processing-Dong, Disambiguation-Hong} heavily rely on linguistic experts to maintain robust dictionaries and complex predefined rules. Typically, such frameworks segment texts into word pieces, disambiguate the pronunciation of the word pieces matched in dictionaries, and apply hand-crafted rules to determine the pronunciation of the undetermined polyphonic characters. However, rule-based approaches often fail due to word pieces having distinctive meanings. For example, "為"您所用~(translate: used "by" you) and "為"您服務~(translate: service "for" you) have the same word piece "為" but different meanings and pronunciations (ㄨㄟ2 and ㄨㄟ4).
\end{CJK*}

On the other hand, the learning-based approaches take contextual information into account to determine the pronunciation of a polyphonic character, such as learned statistical rules \cite{ISCSLP2002-Zirong, ICMLC2008-Huang}, Decision Tree \cite{ICGEC2010-Liu, Computer-Liu}, Maximum Entropy Model \cite{ICASSP2007-Mao, Key-Liu}, and deep learning approaches \cite{ISCSLP2016-Shan,INTERSPEECH2019-Cai,g2pm,INTERSPEECH2020-Zhang,mask-based,INTERSPEECH2019-Dai,INTERSPEECH2019-Yang,INTERSPEECH2021-Shi,pdf}. Among the learning-based approaches, deep learning methods have achieved significant performance for extracting contextual features in polyphone disambiguation. For example, \cite{ISCSLP2016-Shan,INTERSPEECH2019-Cai,g2pm} adopted bidirectional Long Short-Term Memory (BiLSTM) layers to obtain neighboring contextual features for the character of interest. \cite{ISCSLP2016-Shan} and \cite{INTERSPEECH2019-Cai} leveraged additional Part-Of-Speech (POS) tagging and Word2Vec embeddings to obtain more information within a sentence. The sequence-to-sequence model with distant supervision is applied for polyphone disambiguation \cite{INTERSPEECH2020-Zhang}. In \cite{mask-based}, the MASK-BASED approach applied word segmentation and POS tagging within a sentence and restricted the outputs by a weighted-softmax function.

Rather than training prediction models from scratch, recently the pre-trained language model (PLM) has taken advantage of self-supervised learning on vast volumes of unlabeled text data and benefited from downstream tasks after fine-tuning, with examples such as BERT \cite{BERT}. Many studies \cite{g2pm,INTERSPEECH2019-Dai,INTERSPEECH2019-Yang,INTERSPEECH2021-Shi,pdf} have shown that polyphone disambiguation can be improved by leveraging PLM in the polyphone disambiguation problem. For example, \cite{pdf} considered the lattice information to improve performance based on pre-trained BERT.


\begin{figure}[t]
  \centering
  \includegraphics[width=\linewidth]{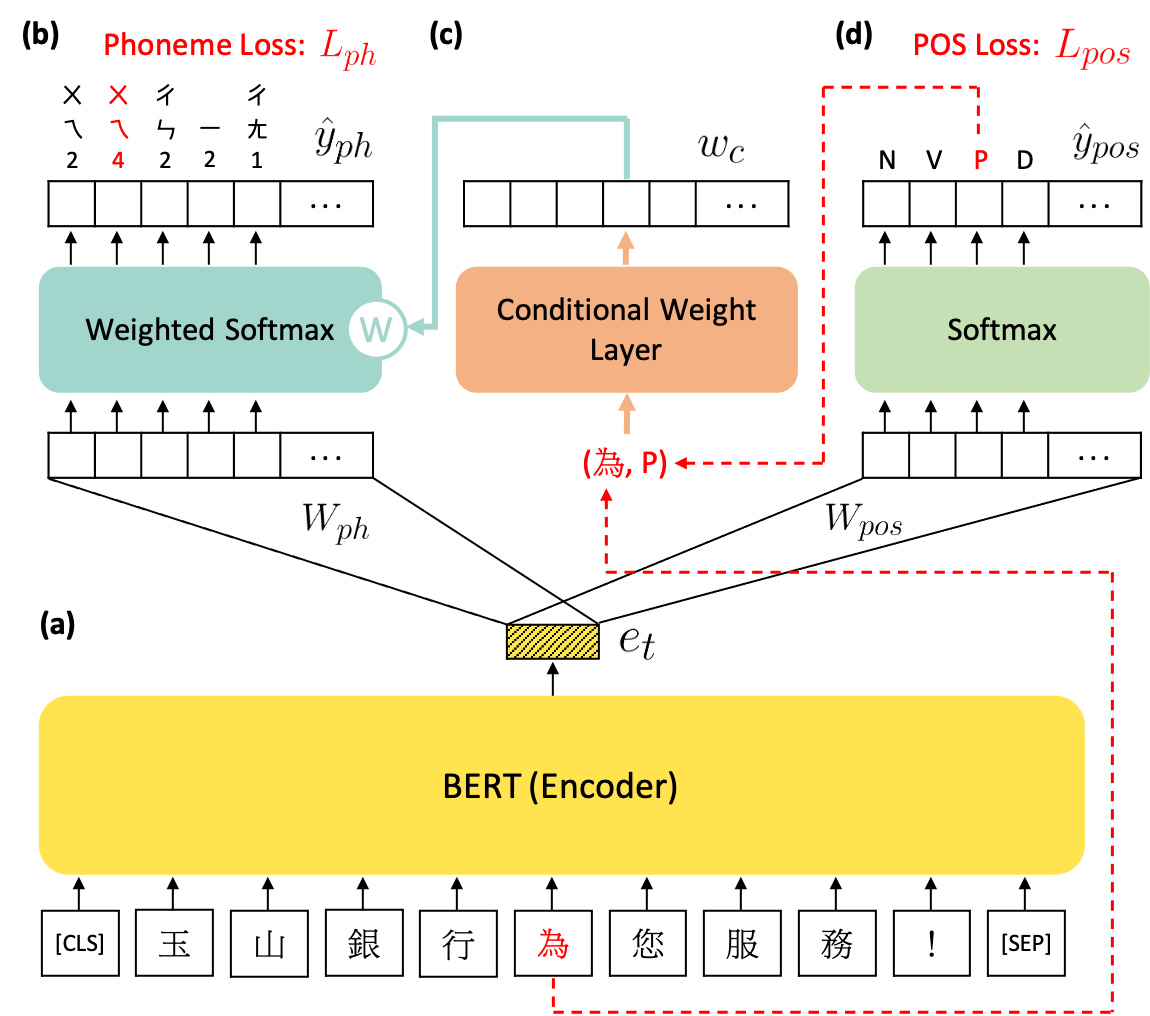}
  \caption{\textbf{The framework of g2pW} includes four components: (a) encoding, (b) phoneme prediction, (c) the conditional weight layer, and (d) POS prediction.}
  \label{fig:g2pW}
\end{figure}

Based on the above-mentioned literature, PLM, restricted output, and extra information from POS tagging have benefited the polyphone disambiguation problem. Inspired by these strategies, we propose a novel approach, called g2p\textbf{W}, which adapts learnable softmax-\textbf{W}eights to condition the outputs of BERT with the polyphonic character of interest and its POS tagging, as shown in Figure~\ref{fig:g2pW}.
Instead of applying the hard mask to softmax-weights as in \cite{mask-based}, our g2pW learns a soft-weighting function for the candidate phonemes. Specifically, our g2pW applies the auxiliary features, such as the character of interest and its POS tagging, to learn an embedding for conditioning the weights in the softmax function. It is worth noting that, unlike previous works \cite{ISCSLP2016-Shan,mask-based}, our g2pW does not need an extra pre-trained POS tagging model. We train our polyphone disambiguation and POS tagging models simultaneously with the unified encoder (BERT). In our framework, we use the predicted POS tag from our joint-trained tagging model as the input for the conditional weight layer. Only a simple text is required for the input of our g2pW, and the simple architecture benefits the inference time. In our experiments, we show that our framework outperforms existing methods on the public CPP dataset \cite{g2pm}. To evaluate the generalization ability of our proposed framework, we also created a new Mandarin polyphone dataset annotated by experts, called MPB (Mandarin Polyphones with Bopomofo). The details will be addressed in Section \ref{exp}. The contributions of this work are summarized as follows:



\begin{itemize}
\item We propose a novel grapheme-to-phoneme model (g2pW) that adapts learnable softmax-weights to condition the outputs of BERT with the polyphonic character of interest and the POS tag extracted from the joint-trained tagging model.
\item Our proposed g2pW outperforms existing methods on the public CPP dataset and achieves 99.08\% accuracy.
\item We released all codes\footnote{\url{https://github.com/GitYCC/g2pW}\label{foot1}}, model weights trained from the MPB dataset\footref{foot1}, and a user-friendly package on PyPi\footnote{\url{https://pypi.org/project/g2pw}}.
\end{itemize}


\section{Proposed method}
\label{method}

As shown in Figure \ref{fig:g2pW}, our proposed g2pW includes four components: (a) encoding, (b) phoneme prediction, (c) the conditional weight layer, and (d) POS prediction. We briefly describe the whole procedure of our framework. For the input of our g2pW, only a raw sentence and the position of the polyphonic character are required. To concentrate the nearby characters of the target polyphonic character, we truncated the raw sentence with a predefined window size $l_{win}$ centered on the target character. The truncated text is also added with special tokens ({\tt [CLS]} and {\tt [SEP]}) before being forwarded to the encoder (BERT). After obtaining the contextual embedding of the target character $e_t$ from the encoder, the feedforward networks $W_{ph}$ and $W_{pos}$ transform $e_t$ into the inputs of the phoneme prediction $e_{ph}$ and the inputs of POS prediction $e_{pos}$. It is worth noting that we extracted the POS tag of the target character directly from our joint-trained POS model. Once the POS tag is obtained, we re-encode the target character and its POS tag to generate the conditional weight $w_c$ from the conditional weight layer to constrain the output of the phoneme prediction model. Detailed descriptions are provided in the following subsections.


\subsection{Phoneme prediction with weighted softmax}
\label{method-1}

As shown in Figure 1(b), we apply a weighted softmax in our phoneme prediction model. Let $n$ be the number of the phoneme labels and suppose that the conditional weights $w_c=\{w_1, w_2, ..., w_{n}\}$ are calculated from conditional weight layer. The weighted softmax is denoted as follows:
\begin{equation}
  \hat{y}_{ph,i} = \frac{w_i\times exp\{[W_{ph}(e_t)]_i\}}{\sum_{j=1}^{n}w_j\times exp\{[W_{ph}(e_t)]_j\}},
  \label{eq:weighted_softmax}
\end{equation}
where $\hat{y}_{ph,i}$ and $[W_{ph}(e_t)]_i$ respectively represent the probability and logit of the $i$-th phoneme label. In our g2pW, the weighted softmax aims to condition the output of the phoneme prediction model by specific prior knowledge, such as what pronunciations we should focus on. We provide a detailed description in the next section. Given the weighted softmax from the conditional weight layer, the phoneme loss can be expressed as follows: 
\begin{equation}
  L_{ph} = \text{CrossEntropy}(\hat{y}_{ph}, y_{ph}),
  \label{eq:phoneme_loss}
\end{equation}
where $y_{ph}$ is the ground truth of the phonemes. 



\subsection{Conditional weight layer}
\label{method-2}

\begin{figure}[t]
  \centering
  \includegraphics[width=1.0\linewidth]{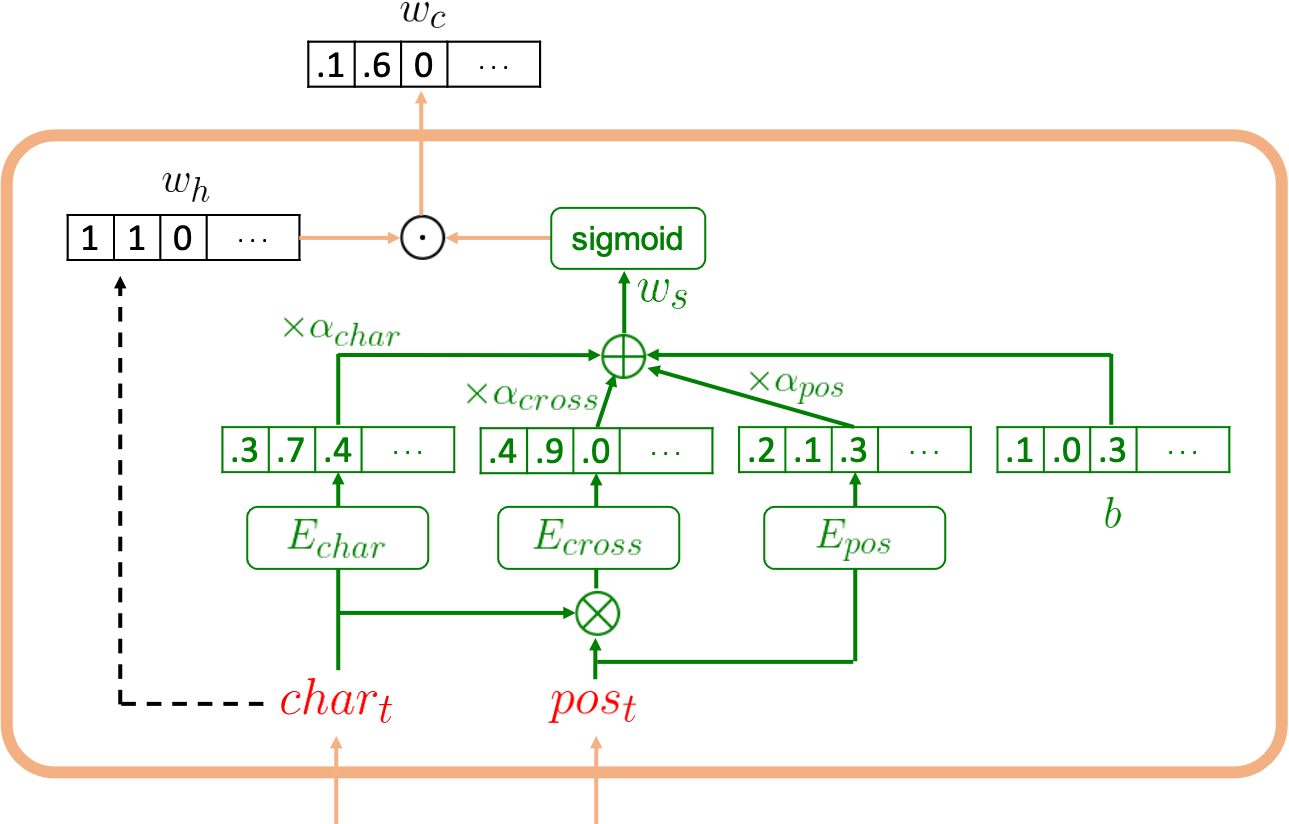}
  \caption{\textbf{The structure of the conditional weight layer}}
  \label{fig:conditional}
\end{figure}

\begin{CJK*}{UTF8}{bsmi}
In our conditional weight layer, we adopt two strategies to obtain the learnable weights for conditioning the output of the phoneme prediction model. One is to reduce the possible label set by giving the target character. For example, there are only two possible pronunciations of 為, and we only need to determine the prediction from these two candidates. To tackle this problem, a weighted softmax with the hard mask/binary mask can be used to restrict outputs \cite{mask-based}. That is, the weights of the non-candidate labels are set to zero as $w_h$ in Figure~\ref{fig:conditional}. Besides the hard mask, we further learn a soft-weighting function for the candidate labels. Specifically, we use the encodings of the target character and its POS tag as the auxiliary features to learn the soft-weighting function as shown in Figure~\ref{fig:g2pW}~(c) and Figure~\ref{fig:conditional}.
\end{CJK*}


Suppose the target character $char_t$ and its POS tag $pos_t$ are given, our soft-weights $w_s$ can be learned as follows:
\begin{equation}
\begin{aligned} 
  w_s&~=~\alpha_{cross}\times E_{cross}(char_t\otimes pos_t)  \\
  \oplus&~\alpha_{char}\times E_{char}(char_t) 
  \oplus \alpha_{pos}\times E_{pos}(pos_t) \oplus b,
 \end{aligned}
  \label{eq:ws}
\end{equation}
where $\otimes$ is a tensor product which maps a pair $(u_i,v_j)$ [$u_i\in u$ and $v_j\in v$] to an element of $u\otimes v$ and $\oplus$ is an element-wise addition. $E_{cross}$, $E_{char}$, and $E_{pos}$ are the learnable linear projection, and $b$ is the learnable bias term. Note that $\alpha_{cross},\alpha_{char},\alpha_{pos}\in \{0,1\}$ are hyper-parameters determined by experiments shown in Table~\ref{table:exp}(b) of Section~\ref{exp-3}. In our experiments, we set $\alpha_{cross}=1$, $\alpha_{char}=1$, and $\alpha_{pos}=0$. Once $w_s$ is determined, the conditional weight $w_c$ is denoted as follows:
\begin{equation}
  w_c = w_h \odot \text{sigmoid}(w_s),
  \label{eq:wc}
\end{equation}
where
\begin{equation}
w_{h,i}=
\begin{cases}
  1, & \text{if candidate phoneme} \\
  0, & \text{otherwise}
\end{cases}
\end{equation}
and $\odot$ is element-wise multiplication.


\subsection{POS prediction}
\label{method-3}
Instead of using an extra pre-trained POS tagging model, we train the POS tagging model and phoneme prediction model simultaneously with the unified encoder (BERT) as shown in Figure~\ref{fig:g2pW}(d). In our joint-trained POS tagging task, we have 11 tags, including  the unknown tag ({\tt UNK}), adjective ({\tt A}), conjunction ({\tt C}), adverb ({\tt D}), interjection ({\tt I}), noun ({\tt N}), preposition ({\tt P}), particle ({\tt T}), verb ({\tt V}), {\tt DE}, and {\tt SHI}. The loss function of our POS tagging model is denoted as follows:
\begin{equation}
  L_{pos} = \text{CrossEntropy}(\hat{y}_{pos}, y_{pos}),
  \label{eq:pos_loss}
\end{equation}
where $y_{pos}$ is the ground truth of POS tagging and $\hat{y}_{pos}$ is the predicted probability of POS tagging. 

In our implementation, we adopt a teacher mode that directly assigns the ground truth $y_{ph}$ to the conditional weight layer during training. At the inference stage, we choose the POS tag with highest probability for the target character. In our proposed g2pW, the total loss can be expressed as follows:
\begin{equation}
  L_{total} = L_{ph} + \beta \times L_{pos},
  \label{eq:total_loss}
\end{equation}
where $\beta$ is the weight that controls the trade-off between $L_{ph}$ and $L_{pos}$.
\\


\section{Experiments}
\label{exp}

In our experiments, we evaluate our g2pW with our proposed MPB dataset and public CPP dataset. We provide the implementation details in Section~\ref{exp-1}. The details of the MPB dataset are presented in Section~\ref{exp-2}. In Section~\ref{exp-3}, we use the MPB dataset to determine the hyperparameters in our g2pW. In Section~\ref{exp-5}, we use the public CPP dataset to benchmark our framework with existing methods.


\subsection{Implementation details}
\label{exp-1}

In our g2pW, the framework learns phoneme prediction, conditional weight layer, and POS prediction on top of the pre-trained encoder (BERT). Huggingface\footnote{\url{https://huggingface.co/bert-base-chinese}} provides the pre-trained BERT model, and the details of BERT are identical to the BERT$_{BASE}$ model (12 layers and 768 hidden sizes) described in \cite{BERT}. Rather than freezing the weights of BERT, we trained our g2pW and fine-tuned BERT simultaneously. For the input of our model, we set the window size $l_{win}$ to truncate the raw text to a subtext with a length of 32 in our experiments. We use the Adam optimizer and set the learning rate to 5e-5. The model is validated after every 200 iterations with a batch size of 256. Once 10,000 iterations is reached, the model with the highest validation accuracy is used for prediction. In our joint-trained POS tagging model, the tags are extracted from Ckiptagger\footnote{\url{https://github.com/ckiplab/ckiptagger}} for training.


\subsection{The MPB dataset}
\label{exp-2}


In our experiments, we first evaluate our g2pW with our in-house Mandarin Polyphone dataset with Bopomofo (MPB). Experts annotated all the pronunciations of the polyphonic characters in the MPB dataset using the \cite{g2pm} format. The MPB dataset includes 2,610,344 sentences and 436 polyphonic characters. Most polyphones are associated with 2 or 3 pronunciations, and only 17 polyphonic characters had fewer than 10 sentences in the dataset. In our experiment, we sample the sentences stratified by the polyphone and split them into the training, development, and test sets with a ratio of 10:1:1. More details are described in Table~\ref{table:stat}.


\begin{table}[t]
\caption{\textbf{Statistics of the MPB dataset}}
\label{table:stat}
\centering
\begin{tabular}{c}
    \subfloat[Dataset division] { 
    \begin{tabular}{lrrrr}
        \toprule
          & \multicolumn{1}{c}{Train} & \multicolumn{1}{c}{Dev.} & \multicolumn{1}{c}{Test} \\
        \midrule
        sentences (\#) &  2,175,097 & 217,597 & 217,650 \\
        polyphonic char.s (\#) & 436 & 433 & 435 \\
        \bottomrule
    \end{tabular}
    }
\end{tabular}
\begin{tabular}{ll}
    \subfloat[Frequency of polyphonic characters according to the numbers of pronunciations] {
    \begin{tabular}{cr}
        \toprule
        pronunc. & \multicolumn{1}{c}{polyphonic} \\
        (\#) & \multicolumn{1}{c}{char.s (\#)}    \\
        \midrule
        2 & 379  \\
        3 & 48  \\
        4 & 5  \\
        5 & 2  \\
        6 & 2 \\
        total & 436 \\
        \bottomrule
    \end{tabular}
    }
    &
    \subfloat[Frequency of polyphonic characters according to the size of sentences] {
    \begin{tabular}{cr}
        \toprule
         sentences & \multicolumn{1}{c}{polyphonic} \\
         (\#) & \multicolumn{1}{c}{char.s (\#)}   \\
        \midrule
        $0-10^1$ & 17  \\
        $10^1-10^2$ & 73  \\
        $10^2-10^3$ & 119  \\
        $10^3-10^4$ & 156  \\
        $10^4-10^5$ & 69  \\
        $10^5-10^6$ & 2  \\
        total & 436 \\
        \bottomrule
    \end{tabular}
    }
\end{tabular}
\end{table}

\begin{table*}[t]
\caption{\textbf{Experimental results on the MPB dataset}}
\label{table:exp}
\centering
\begin{tabular}{cc}
    \subfloat[Evaluation with different $(\alpha_{cross},\alpha_{char},\alpha_{pos})$] { 
    
    \begin{tabular}{ccc|ccc}
        \toprule
        Hard & \multicolumn{1}{c}{Learnable Weight} & POS &  Dev. / Test  \\
        Mask &  {\tiny $(\alpha_{cross},\alpha_{char},\alpha_{pos})$}  &  Joint  &  Acc. (\%) \\
        \midrule
        \checkmark & $(0,1,0)$ & $\beta$=0.1 & 99.62 / 99.62 \\
        \checkmark & $(0,0,1)$ & $\beta$=0.1 & 99.63 / 99.63 \\
        \checkmark & $(0,1,1)$ & $\beta$=0.1 & 99.63 / 99.63 \\
        \checkmark & $(1,0,0)$ & $\beta$=0.1 & 99.63 / 99.62 \\
        \cellcolor{gray!25}\checkmark & \cellcolor{gray!25}$(1,1,0)$ & \cellcolor{gray!25}$\beta$=0.1 & \cellcolor{gray!25}\textbf{99.64} / \textbf{99.64} \\
        \checkmark & $(1,0,1)$ & $\beta$=0.1 & 99.62 / 99.63 \\
        \checkmark & $(1,1,1)$ & $\beta$=0.1 & 99.61 / 99.61 \\
        \bottomrule
    \end{tabular}
    }
    
    &
    \subfloat[Evaluation with different $\beta$] { 
    
    \begin{tabular}{ccl|ccc}
        \toprule
        Hard & \multicolumn{1}{c}{Learnable Weight} & \multicolumn{1}{c|}{POS} & Dev. / Test & Test POS \\
        Mask &  {\tiny $(\alpha_{cross},\alpha_{char},\alpha_{pos})$}  &  \multicolumn{1}{c|}{Joint} &  Acc. (\%) &  Acc. (\%) \\
        \midrule
        \checkmark & $(1,1,0)$ & $\beta$=.01  & 99.62 / 99.60 & 95.73 \\
        \checkmark & $(1,1,0)$ & $\beta$=.02  & \textbf{99.64} / 99.62 & 96.40\\
        \checkmark & $(1,1,0)$ & $\beta$=.05  & 99.62 / 99.62 & 97.14\\
        \cellcolor{gray!25}\checkmark & \cellcolor{gray!25}$(1,1,0)$ & \cellcolor{gray!25}$\beta$=0.1 & \cellcolor{gray!25}\textbf{99.64} / \textbf{99.64} & \cellcolor{gray!25}97.48\\
        \checkmark & $(1,1,0)$ & $\beta$=0.2 & 99.63 / 99.61 & 97.72\\
        \checkmark & $(1,1,0)$ & $\beta$=0.5 & 99.62 / 99.61 & 97.92\\
        \checkmark & $(1,1,0)$ & $\beta$=1.0 & 99.60 / 99.59 & 97.91\\
        \bottomrule
    \end{tabular}
    }
    
\end{tabular}
\begin{tabular}{c}
\subfloat[Contribution investigation within our framework] { 

\begin{tabular}{c|ccc|cccc}
    \toprule
    System & Hard  & Learnable Weight  & POS  & Test Acc.  & Test Averaged Accuracy  & Test POS  \\
    & Mask & $(\alpha_{cross},\alpha_{char},\alpha_{pos})$ & Joint & (\%) & by Characters (\%) & Acc. (\%) \\
    \midrule
    \cellcolor{gray!25}g2pW & \cellcolor{gray!25}\checkmark & \cellcolor{gray!25}$(1,1,0)$ & \cellcolor{gray!25}$\beta$=0.1 & \cellcolor{gray!25}\textbf{99.64} & \cellcolor{gray!25}\textbf{95.25} & \cellcolor{gray!25}97.48 \\
     & \checkmark & $\times$ & $\beta$=0.1  & 99.61 & 94.47 & 97.46 \\
      & \checkmark & $\times$ & $\times$  & 99.60 & 94.63 & - \\
    baseline & $\times$ & $\times$ & $\times$  & 99.50 & 91.18 & - \\
    \bottomrule
\end{tabular}
}
\end{tabular}
\end{table*}

\begin{table}[t]
  \caption{\textbf{Benchmarks on the CPP dataset}}
  \label{table:cpp}
  \centering
  \begin{tabular}{ lcc }
    \toprule
    \multicolumn{1}{c}{System} & Year & Test Acc. (\%) \\
    \midrule
    g2pM (BiLSTM) \cite{g2pm} & 2020 & 97.31            \\
    Distant supervision \cite{INTERSPEECH2020-Zhang} & 2020 & 97.51              \\
    MASK-BASED \cite{mask-based}  & 2020 & 97.68       \\
    g2pM (BERT) \cite{g2pm}  & 2020 &  97.85           \\
    BERT with LSTM \cite{pdf}  & 2021 & 98.04       \\
    PDF (with BERT) \cite{pdf} & 2021 & 98.83          \\
    g2pW &  &   \textbf{99.08}        \\
    \bottomrule
  \end{tabular}
\end{table}

\subsection{Experiments on the MPB dataset}
\label{exp-3}
In our g2pW, $(\alpha_{cross},\alpha_{char},\alpha_{pos})$ in \eqref{eq:ws} and $\beta$ in \eqref{eq:total_loss} are two crucial hyper-parameters. In this experiment, we evaluate our g2pW with the MPB dataset and determine these hyper-parameters.\\


\noindent \textbf{Evaluation with different $(\alpha_{cross},\alpha_{char},\alpha_{pos})$.} To determine the optimal combination of $(\alpha_{cross},\alpha_{char},\alpha_{pos})$, we intuitively assigned $\beta$ as 0.1 since the main objective of our task is phoneme prediction. By fixing the value of $\beta$, we present results of all the combinations of $(\alpha_{cross},\alpha_{char},\alpha_{pos})$ in Table~\ref{table:exp}(a) and show that $(\alpha_{cross},\alpha_{char},\alpha_{pos})=(1,1,0)$ achieves the highest accuracy. Therefore, we use this combination in our g2pW.\\


\noindent \textbf{Evaluation with different $\beta$.} Once the optimal $(\alpha_{cross},\alpha_{char},\alpha_{pos})$ is determined, we further search for a proper $\beta$.  As mentioned in Section~\ref{method-3}, $\beta$ is the weight that controls the trade-off between $L_{ph}$ and $L_{pos}$. This experiment searches the optimal $\beta$ between 0.01 to 1 since the main objective is phoneme prediction. As shown in Table~\ref{table:exp}(b), a larger $\beta$ achieves more accurate POS tagging. However, over-emphasizing the loss of the POS tagging task will sacrifice the performance of phoneme prediction. In Table~\ref{table:exp}(b), according to the highest accuracy, we chose $\beta=0.1$ in our g2pW.


\noindent \textbf{Contribution investigation within our framework.} In Table~\ref{table:exp}(c), we first compare our g2pW with the baseline in which the pure fine-tuned BERT is used. Our g2pW provides an accuracy improvement of 0.14\% from the baseline. On the other hand, as shown in Table~\ref{table:stat}(c), the accuracy of a particular character with larger cardinality will dominate the accuracies of other characters with smaller cardinality. Thus, we also reported the averaged accuracy by characters: $\frac{1}{C}\sum_c \text{accuracy}(\{(\hat{y}_{ph}, y_{ph})\mid char_t = c\})$ where $C$ is the number of all polyphonic characters. As shown in Table~\ref{table:exp}(c), our g2pW provides a significant improvement of  4.07\% to the test averaged accuracy by character compared with the baseline.

As shown in Figure~\ref{fig:g2pW}, our g2pW is composed of several critical components, such as the hard mask, conditional weight layer, and joint-trained POS tagging task. According to the results from the baseline model, removing all these components will sacrifice performance. In the following experiments, we investigate the discrete contributions of these components. The first experiment removeds the conditional weight layer and directly uses the hard mask in the weighted softmax. Compared to g2pW, removing the conditional weight layer will decrease accuracy, as shown in the second row of Table~\ref{table:exp}(c). The second experiment removes the POS joint-training tagging task and obtains worse performance, as shown in the third row of Table~\ref{table:exp}(c). The third and final experiment removes the remaining hard mask in our framework (i.e., the baseline model). We set $w_h$ in \eqref{eq:wc} by a vector filled with the scalar value 1. From the result of the baseline model, removing the hard mask has a significantly negative impact on performance. It is worth noting that our g2pW still can provide further improvements on top of the significant performance achieved by using the hard mask.



\subsection{Public benchmark}
\label{exp-5}

In our final experiment, we compare our proposed g2pW with existing approaches by using the public CPP dataset \cite{g2pm}. The implementation details and experimental settings are the same as the experiments of the MPB dataset described in Section~\ref{exp-1}. From the results of Table~\ref{table:cpp}, our g2pW achieves 99.08\% test accuracy and outperforms the existing approaches. Compared with the state-of-the-art method, PDF (with BERT) \cite{pdf}, our g2pW improves 0.25\% testing accuracy. It is worth noting that PDF (with BERT) requires an extra dictionary to extract the lattice information while only raw sentences are required as the input in our g2pW.


\section{Conclusion}

We propose a novel grapheme-to-phoneme model (g2pW) that adapts learnable softmax-weights to condition the outputs of BERT. In our experiments, we show the conditional weighted softmax conditioned with the polyphonic character of interest and its POS tagging improves performance of polyphone disambiguation. From the results, we also show that our g2pW outperforms existing methods on the public CPP dataset.

\section{Acknowledgements}

This work was supported in part by the E.SUN Financial Holding CO., LTD. of Taiwan and the Ministry of Science and Technology of Taiwan under Grants MOST 108-2221-E-017-008-MY3.

\bibliographystyle{IEEEtran}

\bibliography{mybib}

\begin{thebibliography}{10}
\providecommand{\url}[1]{#1}
\csname url@samestyle\endcsname
\providecommand{\newblock}{\relax}
\providecommand{\bibinfo}[2]{#2}
\providecommand{\BIBentrySTDinterwordspacing}{\spaceskip=0pt\relax}
\providecommand{\BIBentryALTinterwordstretchfactor}{4}
\providecommand{\BIBentryALTinterwordspacing}{\spaceskip=\fontdimen2\font plus
\BIBentryALTinterwordstretchfactor\fontdimen3\font minus
  \fontdimen4\font\relax}
\providecommand{\BIBforeignlanguage}[2]{{%
\expandafter\ifx\csname l@#1\endcsname\relax
\typeout{** WARNING: IEEEtran.bst: No hyphenation pattern has been}%
\typeout{** loaded for the language `#1'. Using the pattern for}%
\typeout{** the default language instead.}%
\else
\language=\csname l@#1\endcsname
\fi
#2}}
\providecommand{\BIBdecl}{\relax}
\BIBdecl

\bibitem{Processing-Dong}
H.~Dong, J.~Tao, and B.~Xu, ``Processing of polyphone character in chinese tts
  system,'' \emph{Chinese Information}, no.~01, pp. 33--36.

\bibitem{Disambiguation-Hong}
Z.~Hong, Y.~Jiangsheng, Z.~Weidong, and Y.~Shiwen, ``Disambiguation of chinese
  polyphonic characters,'' in \emph{The First International Workshop on
  MultiMedia Annotation (MMA)}, 2001.

\bibitem{ISCSLP2002-Zirong}
Z.~Zirong, C.~Min, and C.~Eric, ``An efficient way to learn rules for
  grapheme-to-phoneme conversion in chinese,'' in \emph{International Symposium
  on Chinese Spoken Language Processing (ISCSLP)}, 2002.

\bibitem{ICMLC2008-Huang}
F.-L. Huang, ``Disambiguating effectively chinese polyphonic ambiguity based on
  unify approach,'' in \emph{ICMLC}, 2008.

\bibitem{ICGEC2010-Liu}
J.~Liu, W.~Qu, X.~Tang, Y.~Zhang, and Y.~Sun, ``Polyphonic word disambiguation
  with machine learning approaches,'' in \emph{International Conference on
  Genetic and Evolutionary Computing (ICGEC)}, 2010.

\bibitem{Computer-Liu}
F.~Liu and Y.~Zhou, ``Polyphone disambiguation based on tree-guided tbl,''
  \emph{Computer Engineering and Applications}, vol.~47, no.~12, pp. 137--140,
  2011.

\bibitem{ICASSP2007-Mao}
X.~Mao, Y.~Dong, J.~Han, D.~Huang, and H.~Wang, ``Inequality maximum entropy
  classifier with character features for polyphone disambiguation in mandarin
  tts systems,'' \emph{2007 IEEE International Conference on Acoustics, Speech
  and Signal Processing - ICASSP '07}, pp. IV--705--IV--708, 2007.

\bibitem{Key-Liu}
F.~Z. Liu and Y.~Zhou, ``Polyphone disambiguation based on maximum entropy
  model in mandarin grapheme-to-phoneme conversion,'' \emph{Key Engineering
  Materials}, vol. 480--481, pp. 1043--1048, 2011.

\bibitem{ISCSLP2016-Shan}
C.~Shan, L.~Xie, and K.~Yao, ``A bi-directional lstm approach for polyphone
  disambiguation in mandarin chinese,'' \emph{2016 10th International Symposium
  on Chinese Spoken Language Processing (ISCSLP)}, pp. 1--5, 2016.

\bibitem{INTERSPEECH2019-Cai}
Z.~Cai, Y.~Yang, C.~Zhang, X.~Qin, and M.~Li, ``Polyphone disambiguation for
  mandarin chinese using conditional neural network with multi-level embedding
  features,'' in \emph{INTERSPEECH}, 2019.

\bibitem{g2pm}
K.~Park and S.~Lee, ``g2pm: A neural grapheme-to-phoneme conversion package for
  mandarin chi- nese based on a new open benchmark dataset,'' in
  \emph{INTERSPEECH}, 2020.

\bibitem{INTERSPEECH2020-Zhang}
J.~Zhang, Y.~Zhao, J.~Zhu, and J.Xiao, ``Distant supervision for polyphone
  disambiguation in mandarin chinese,'' in \emph{INTERSPEECH}, 2020.

\bibitem{mask-based}
H.~Zhang, H.~Pan, and X.~Li, ``A mask-based model for mandarin chinese
  polyphone disambiguation,'' in \emph{INTERSPEECH}, 2020.

\bibitem{INTERSPEECH2019-Dai}
D.~Dai, Z.~Wu, S.~Kang, X.~Wu, J.~Jia, D.~Su, D.~Yu, and H.Meng,
  ``Disambiguation of chinese polyphones in an end-to-end framework with
  semantic features extracted by pre-trained bert,'' in \emph{INTERSPEECH},
  2019.

\bibitem{INTERSPEECH2019-Yang}
B.~Yang, J.~Zhong, and S.~Liu, ``Pre-trained text representations for improving
  front-end text processing in mandarin text-to-speech synthesis,'' in
  \emph{INTERSPEECH}, 2019.

\bibitem{INTERSPEECH2021-Shi}
Y.~Shi, C.~Wang, Y.~Chen, and B.~Wang, ``Polyphone disambiguition in mandarin
  chinese with semi-supervised learning,'' in \emph{INTERSPEECH}, 2021.

\bibitem{pdf}
H.~Zhang, ``Pdf: Polyphone disambiguation in chinese by using flat,'' in
  \emph{INTERSPEECH}, 2021.

\bibitem{BERT}
J.~Devlin, M.-W. Chang, K.~Lee, and K.~Toutanova, ``Bert: Pre-training of deep
  bidirectional transformers for language understanding,'' in \emph{arXiv
  preprint arXiv:1810.04805}, 2018.

\end{thebibliography}


\end{document}